# Radiology Report Generation with Layer-Wise Anatomical Attention

Emmanuel D. Muñiz-De-León[1], Jorge A. Rosales-de-Golferichs[1], Ana S. Muñoz-Rodríguez[1],
Alejandro I. Trejo-Castro[2], Eduardo de Avila-Armenta[3],
Antonio Martínez-Torteya[1][*]

[1]School of Engineering and Technology, Universidad de Monterrey, San Pedro Garza García, Mexico.
[2]School of Medicine and Health Sciences, Tecnológico de Monterrey, Monterrey, Mexico
[3]School of Engineering and Sciences, Tecnológico de Monterrey, Monterrey, Mexico

*Abstract*—Automatic radiology report generation is a promising application of multimodal deep learning, aiming to reduce reporting workload and improve consistency. However, current state-of-the-art (SOTA) systems—such as Multimodal AI for Radiology Applications (MAIRA-2) and Medical Pathways Language Model-Multimodal (MedPaLM-M)—depend on large-scale multimodal training, clinical metadata, and multiple imaging views, making them resource-intensive and inaccessible for most settings.

We introduce a compact image-to-text architecture that generates the *Findings* section of chest X-ray reports from a single frontal image. The model combines a frozen Self-Distillation with No Labels v3 (DINOv3) Vision Transformer (ViT) encoder with a Generative Pre-trained Transformer 2 (GPT-2) decoder enhanced by layer-wise anatomical attention. This mechanism integrates lung and heart segmentation masks through hierarchical Gaussian smoothing, biasing attention toward clinically relevant regions without adding trainable parameters.

Evaluated on the official Medical Information Mart for Intensive Care-Chest X-ray (MIMIC-CXR) dataset using Chest Radiograph Expert (CheXpert) and Radiology Graph (RadGraph) metrics, our approach achieved substantial gains: CheXpert Macro-F1 for five key pathologies increased by 168% (0.083 → 0.238) and Micro-F1 by 146% (0.137 → 0.337), while broader performance across 14 observations improved by 86% (0.170 → 0.316). Structural coherence also improved, with RadGraph F1 rising by 9.7%. Despite its small size and purely image-conditioned design, the model demonstrates that decoder-level anatomical guidance improves spatial grounding and enhances coherence in clinically relevant regions.

The source code is publicly available at: https://github.com/devMuniz02/UDEM-CXR-Reporting-Thesis-2025.

## I. INTRODUCTION

Radiology report generation has emerged as a key challenge at the intersection of medical imaging and natural language processing [1]. Chest X-rays (CXRs) remain one of the most widely used diagnostic modalities worldwide [2], yet producing radiology reports is time-consuming and prone to variability driven by workload, expertise, and institutional practice differences [3]. This challenge emphasizes the need for systems that can enhance efficiency and consistency [4].

Automated report generation systems offer the potential to improve consistency, reduce the reporting burden [3], and enhance access to expert level interpretation, particularly in resource limited environments [1]. The evolving importance of Artificial Intelligence (AI) in medical trainee education highlights its future role in mitigating these pressures and training the next generation of radiologists [5]. However, despite the development of generative AI for interpretation [1][4], generating reports that are clinically faithful, anatomically grounded, and diagnostically precise continues to be a difficult, unsolved problem [1].

Early approaches relied on template-based or retrieval-based methods, which lacked the generative flexibility required for nuanced diagnostic reasoning. The advent of transformer-based multimodal architectures led to substantial progress: models such as Radiology Report Generation (R2Gen) [6], Multimodal AI for Radiology Applications (MAIRA) models MAIRA-1 [7], MAIRA-2 [8], and Medical Pathways Language Model-Multimodal (MedPaLM-M) [9] leverage large-scale pretraining, multimodal alignment, and medical grounding to improve report quality. Yet these systems typically require significant computational resources, complex pipelines, and auxiliary inputs such as structured labels, clinical metadata, or ontology-based reasoning modules [7][8][9].

Parallel to these advances, a growing line of research emphasizes the importance of anatomical grounding in medical vision-language models. Methods such as Anatomy-XNet [10], MAIRA-SEG [11], the MXA attention block [12], and transformer-based structural modules like UNeST [13] demonstrate that incorporating anatomical priors improves spatial alignment, reduces hallucinations, and enhances interpretability. However, most existing approaches inject anatomical information into the visual encoder or visual attention blocks [12][13], leaving the decoder (the component responsible for generating clinical language) largely unaware of explicit spatial structure.

In contrast, this work introduces a decoder-centered anatomical grounding mechanism. We propose a compact multimodal architecture that combines a frozen Self-

---

[*]Corresponding Author: antonio.martinez@udem.edu

Distillation with No Labels v3 (DINOv3) Vision Transformer (ViT) Small (16x16 patch) encoder [14] (22M parameters), a lightweight linear projection adapter, and a modified GPT-2 decoder (124M parameters) [15] enriched with layer-wise anatomy-aware attention maps derived from lung and heart segmentation masks. This design allows the language model to integrate spatial anatomical cues directly during autoregressive generation, without modifying the visual encoder or increasing the number of trainable parameters. The system is purely image-conditioned, receiving no prompts, metadata, or auxiliary textual inputs.

The model is trained and evaluated using the frontal images from the MIMIC Chest X-ray JPG (MIMIC-CXR-JPG) dataset [16], a version of the MIMIC database provided in JPG format for increased computational efficiency. We specifically filter the dataset to include only those studies containing a "Findings" section, ensuring the model learns from the primary descriptive portion of the radiology report.

Clinical quality is assessed using pathology F1 scores based on labels extracted by CheXbert [17], a BERT-based model from the original CheXpert dataset [2], covering both the full 14-pathology set and a specific 5-class subset of clinically significant findings in the text. To evaluate structural accuracy, we utilize RadGraph F1 metrics derived from the RadGraph model [18], which parses clinical entities and their relations into a knowledge graph. These benchmarks enable a rigorous comparison with State-of-the-Art (SOTA) systems. Although the proposed model does not reach the performance scale of MAIRA-2 or MedPaLM-M, results demonstrate that integrating anatomical attention directly into the decoder improves spatial grounding and enhances coherence in clinically relevant regions.

Overall, these findings suggest that embedding anatomical structure within the language generation mechanism rather than limiting it to the visual encoding stage offers a promising direction for developing compact, efficient, and clinically aligned radiology report generation models.

## II. METHODOLOGY

To integrate the anatomical priors illustrated in Figure 1, the methodology utilizes a multi-stage approach for automated clinical documentation. As shown in the architectural diagram, the process begins with the input Image, which is processed through a bifurcated pipeline. The first branch employs a Vision Encoder and a specialized Adapter to translate raw pixels into aligned visual embeddings. Simultaneously, the second branch utilizes Segmentation Models dedicated to the LUNG and HEART regions to generate localized Attention Masks. Finally, as depicted in Figure 1, these anatomical masks and visual features are fused within the Language Model to produce a coherent Generated Report that is spatially grounded in specific regions of interest.

### A. Task definition

The objective of this work is to automatically generate the *Findings* section of a radiology report from a single frontal chest X-ray (CXR). The task is defined as a pure image-to-text generation problem, where the model must produce a clinically coherent description of radiological findings using only the image as input. No prompts, clinical histories, demographic information, or auxiliary textual conditioning are provided.

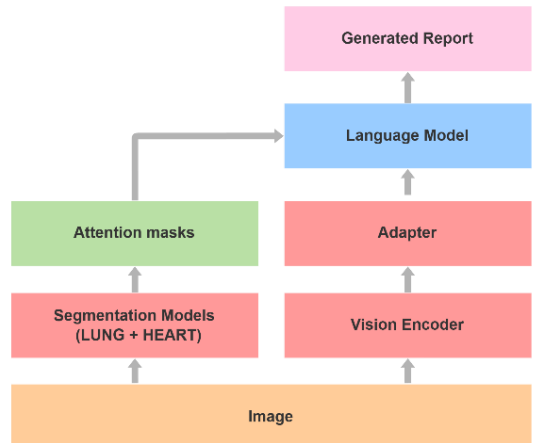

**Fig. 1.** Proposed Layer-wise Anatomical Attention Bias Model.

This task formulation follows the standard adopted in recent radiology report generation benchmarks, where only the *Findings* section is evaluated because it directly describes observations grounded in the image [8].

Given an input CXR and its associated report, the goal of the model is to learn the mapping from the image to the sequence of tokens corresponding to the *Findings* section. During training, the model receives the ground-truth text through teacher forcing, where the decoder predicts the next token based on the actual preceding tokens from the reference report. In contrast, at inference time, the generation is fully autoregressive: the system produces one token at a time, with each prediction conditioned solely on the previously generated tokens and the visual representation of the image. This setting allows the evaluation of image-conditioned generative ability without any external language priors.

This formulation differs from that adopted by large-scale multimodal systems such as MAIRA-2 [8] or MedPaLM-M [9], which typically introduce additional conditioning signals such as impressions, structured labels, multi-stage grounding, or clinical histories. Instead, our approach isolates the core task of radiology report generation to assess the effectiveness of decoder-level anatomical attention within a compact architecture.

The model is trained and evaluated using the official splits of MIMIC-CXR-JPG [16], following established protocols in clinical NLP and radiology report generation, as detailed in the next section. This ensures comparability with prior work and focuses evaluation strictly on the medical content expressed in the Findings, measured through clinically oriented metrics such as CheXpert F1 [17], which evaluates a model's accuracy in classifying the presence of specific thoracic pathologies, and RadGraph F1 metrics [18], which assess the quality of structured information extraction by verifying the correctness of clinical entities and their relations. By utilizing these

*B. Data*

The development and evaluation of the automatic radiological report generation model were carried out using the MIMIC-CXR-JPG dataset, one of the largest and most representative collections of chest radiographs with free-text clinical reports [16][19][20]. The dataset contains 377,110 DICOM images corresponding to 227,835 studies obtained from 65,379 patients, sourced from the Beth Israel Deaconess Medical Center and publicly distributed in an anonymized manner through PhysioNet.

For this work, only frontal anteroposterior (AP) or posteroanterior (PA) radiographs were selected, following criteria consistent with recent report generation studies such as MAIRA-1 [7] and MAIRA-2 [8], which emphasize that these views are the most frequent and clinically relevant for automated detection of thoracic findings. Furthermore, all studies whose reports did not include the *Findings* section were eliminated, as the proposed model was trained exclusively to generate this section.

The precise extraction of the *Findings* sections was performed using the official script provided by the MIT Laboratory for Computational Physiology [21]. This script looks for the standard headers that appear in radiology reports and uses them to separate the document into its main parts. By automatically identifying where each section begins, it allows us to isolate the Findings text in a consistent and reliable way, even when reports vary in formatting or writing style.

This procedure guarantees that the text used for training preserves the original semantics described by the radiologists, avoiding common inconsistencies (e.g., misspellings such as "findnings" instead of "findings") from manual or heuristic segmentations, and is the same method adopted by advanced multimodal works like MAIRA-2 [8].

The official MIMIC-CXR-JPG dataset is divided into three predefined splits to ensure consistency and comparability across studies. The test set includes patients with at least one report that was manually labeled, the validation set consists of a random sample of 500 patients, and the remaining patients form the training set [16]. After applying the filtering criteria described earlier, such as selecting only frontal views and reports that contain a structured *Findings* section, the number of patients, studies, and images was reduced from the original distribution. The final dataset composition after filtering is presented in Table 1

TABLE 1. FINAL DISTRIBUTION OF STUDIES IN MIMIC-CXR-JPG FOR REPORT GENERATION.

| Split | Patients | Studio IDs | DICOMs |
|---|---|---|---|
| Train | 58,689 | 146,135 | 162,969 |
| Valid | 448 | 1,151 | 1,286 |
| Test | 285 | 2,210 | 2,461 |

For the final segmentation stage, specialized image sets with corresponding anatomical lung and heart masks were sourced from multiple institutional and public repositories to train 2 dedicated segmentation models.

To train the auxiliary anatomical segmentation module for the lungs, specific datasets with pixel-level annotations were incorporated:

- Base Training: The model's initial training utilized the Chest X-ray Dataset with Lung Segmentation [22] from PhysioNet, establishing a foundational understanding of lung boundaries.
- Fine-Tuning and Validation: To maximize precision and accuracy, the model was subsequently fine-tuned and validated using high-quality, professionally annotated lung masks from three distinct collections: the Montgomery County X-ray Set [23], the Shenzhen Hospital X-ray Set [23], and the Darwin dataset [24]. These datasets served as the gold standard for refining boundary accuracy.

For cardiac silhouette segmentation, the CheXmask database [25] was employed, which provides 657,566 high-quality anatomical segmentation masks specifically tailored for chest radiographs. The model's generalization capability was evaluated on 247 chest radiographs from the independent Japanese Society of Radiological Technology (JSRT) dataset [26] to confirm its robustness beyond the training distribution.

*C. Preprocessing*

Preprocessing was essential to make sure that the images and texts could be used consistently by the multimodal model. In the case of images, all radiographs were resized to a standard resolution of $512 \times 512$ pixels using LANCZOS interpolation, a method widely employed in computer vision due to its ability to preserve anatomical details and high-frequency edges in medical images.

Given that the DINOv3 [14] architecture was pre-trained on the ImageNet [27] dataset using images in RGB format, the typical single-channel grayscale of the radiographs was replicated to create a synthetic three-channel image compatible with the visual encoder. This procedure is common in recent medical multimodal models like MAIRA-1 [7] and MAIRA-2 [8], which also utilize ViT based encoders pre-trained in RGB domains to capture high-level features.

Subsequently, the pixel values were normalized following the standard ImageNet statistics [27], using a mean ($\mu$) of $[0.485, 0.456, 0.406]$ and a standard deviation ($\sigma$) of $[0.229, 0.224, 0.225]$ for the RGB channels, respectively. This normalization guarantees that the distribution of radiograph intensities is compatible with that used during the encoder's pre-training, which favors visual transfer and training stability.

For the text, a cleaning process was applied to each report that eliminated unnecessary line breaks and excessive or irregular spaces, while keeping the original semantic content intact. This process was also essential for correcting word fragmentation caused by fixed-width formatting in the original reports (e.g., reconstructing terms such as "can-cer" into "cancer" when a word was split across two lines). The result was a homogeneous textual sequence, ideal for tokenization and subsequent input to the language model via teacher forcing, following recommended practices in modern radiological report generation systems [8].

*D. Segmentation*

Lung and heart segmentation was performed using two specialized models, each built on the same hybrid encoder–decoder framework but trained independently to maximize anatomical precision for their respective structures. Both models employed a frozen DINOv3 ConvNeXt-Small encoder (50M parameters) as the visual backbone. The encoder produced 384-dimensional feature maps that were projected to 512 channels through a $1 \times 1$ convolutional adapter to ensure compatibility with a lightweight U-Net–style decoder. This decoder comprised a sequence of convolutional and transposed-convolution blocks designed to progressively upsample the spatial representations and reconstruct high-resolution binary masks.

*1) Lung Segmentation*

To obtain robust lung masks across diverse radiographic domains, a two-stage transfer learning strategy was used. First, the model was pre-trained on the large-scale Lung Segmentation Dataset [22], enabling it to capture general pulmonary morphology and intensity patterns. Subsequently, the network was fine-tuned on expert annotated datasets with high-quality ground truth: Montgomery [23], Shenzhen [23], and Darwin [24]. This refinement stage improved the model's ability to delineate pulmonary boundaries, hilum transitions, and costophrenic recesses.

*2) Heart Segmentation*

The segmentation of the cardiac silhouette was carried out using the same hybrid architecture but trained directly on the CheXmask dataset [25], which provides detailed pixel-level annotations for the heart region. Unlike the lungs (whose boundaries vary significantly across datasets) the heart model benefited from training on a single high-quality annotated source and the model generalization was assessed on the JSRT dataset [26].

*3) Final Segmentation Model*

Following the independent training, the two specialized segmentation models were merged into a single model for inference. This consolidated model efficiently executes a forward pass on the input image to simultaneously generate both the final lung mask and the heart silhouette mask.

*E. Anatomic Attention per layer*

The layer-wise anatomical attention mechanism constitutes one of the core components of this work. Recent studies have consistently demonstrated that the explicit incorporation of anatomical or structural information into Transformer-based architectures can substantially improve the model's capacity for localization, description, and reasoning about clinically relevant regions in medical images. This principle has been explored in three main lines: (1) structural attention within the visual encoder, such as in UNeST [13]; (2) specialized attention blocks for radiographs like MXA [12]; and (3) explicit anatomical attention mechanisms based on segmented masks, such as in Anatomy-XNet [10] and MAIRA-SEG [11].

However, most of these approaches incorporate anatomical information during visual encoding or spatial pre-processing. In contrast to these works, where segmentation is typically used inside the encoder or as part of the input prompt, our approach integrates it directly into the textual decoder, ensuring that every layer of the model is explicitly informed by the relevant anatomy during the autoregressive generation of the report.

*1) Mask construction*

The process begins with a segmentation model that produces binary lung and heart masks at a resolution of $32 \times 32$, which matches the spatial layout of the DINOv3 visual tokens ($N = 1024$). These masks (Figure 2a–2b) are merged into a single fused anatomical mask (Figure 2c). From this fused mask, a set of smoothed masks $\{M_l\}_{l=1}^{L}$, where $M_l \in \mathbb{R}^{32 \times 32}$, is generated by applying separable Gaussian filters with a decreasing standard deviation $\sigma_l \downarrow$. This produces a hierarchical "general-to-specific" progression: early decoder layers use broad, diffuse masks that capture coarse anatomical context, whereas later layers use sharper masks that focus attention on the lung and heart regions (Figure 2d). The size of the one-dimensional Gaussian kernel for each layer is defined as

$$k_l = k_{\text{base}} + (L - (l - 1)) \cdot k_{\text{incr}},$$

with:

$$k_{\text{base}} = 3, k_{\text{incr}} = 2,$$

and the standard deviation approximated by:

$$\sigma_l \approx \frac{k_l - 1}{6}.$$

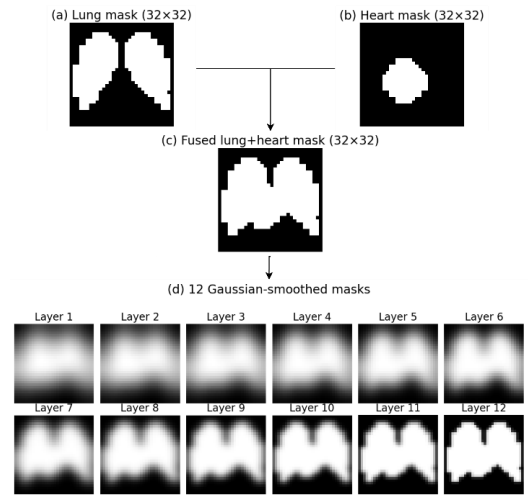

**Fig. 2.** Hierarchical Construction and Smoothing of Layer-Wise Anatomical Attention Masks. (a) Lung Mask (32 x 32); (b) Heart Mask (32 x 32); (c) Fused lung + heart Mask (32 x 32); (d) 12 Gaussian-smoothed masks.

Thus, layer $l = 1$ uses the largest kernel ($27 \times 27$), yielding the most diffuse mask, while layer $l = 12$ uses the smallest kernel ($5 \times 5$), yielding the most focused mask. Algorithm 1 formalizes this hierarchical construction process.

**Algorithm 1** Hierarchical Construction of Layer-Wise Anatomical Masks

1: **Input:** chest radiograph $x$; segmentation model $S$; decoder depth $L$; base kernel $k_{\text{base}}$; increment $k_{\text{incr}}$
2: **Output:** smoothed masks $\{M_l\}_{l=1}^{L}$, $M_l \in \mathbb{R}^{32 \times 32}$
3: $(B^{\text{lung}}, B^{\text{heart}}) \leftarrow S(x)$  ▷ Binary masks at $32 \times 32$
4: $B \leftarrow B^{\text{lung}} \vee B^{\text{heart}}$  ▷ Fused lung–heart mask
5: **for** $l = 1$ to $L$ **do**
6:   $k_l \leftarrow k_{\text{base}} + (L - (l - 1)) \cdot k_{\text{incr}}$
7:   $\sigma_l \leftarrow (k_l - 1)/6$
8:   construct 1-D Gaussian kernel $g_l$ of size $k_l$ and std. $\sigma_l$
9:   $M_l \leftarrow g_l * (g_l^\top * B)$  ▷ Separable Gaussian smoothing
10:   optionally normalize $M_l$ to $[0, 1]$
11: **end for**

*2) Anatomic mask integration*

The flattened anatomical vector $m_l \in \mathbb{R}^N$ is then replicated row-wise according to the current report length $T_{\text{rep}}$ to create the tile matrix

$$T_l = \mathbf{1}_{T_{\text{rep}}} m_l^\top, T_l \in \mathbb{R}^{T_{\text{rep}} \times N}, N = 1024,$$

where $\mathbf{1}_{T_{\text{rep}}}$ is a column vector of ones. This ensures that every token in the report has access to the same anatomical prior.

To restrict this matrix to the causal attention pattern, the tile matrix is element-wise multiplied with the causal mask $C$:

$$B_l = T_l \odot C.$$

The resulting anatomical bias matrix $B_l$ is then added to the raw attention logits before the Softmax,

$$A'_l = A_l + B_l,$$

biasing the decoder toward attending more heavily to tokens associated with anatomically relevant regions, the effect can be seen in Figure 3. This logit-shaping mechanism is conceptually similar to ALiBi-style biasing [28], but instead of positional distance, it uses anatomically derived masks. Because the masks are generated from segmentation and fixed Gaussian filters, the method introduces no additional trainable parameters. Algorithm 2 summarizes this integration.

**Algorithm 2** Layer-Wise Anatomical Attention Bias Integration

1: **Input:** smoothed masks $\{M_l\}_{l=1}^{L}$; cross-attention logits $\{A_l\}_{l=1}^{L}$; causal/padding mask $C \in \{0,1\}^{T_{\text{rep}} \times N}$; visual token count $N = 32 \times 32$
2: **Output:** biased logits $\{A'_l\}_{l=1}^{L}$
3: **for** $l = 1$ to $L$ **do**
4:   $m_l \leftarrow \text{Flatten}(M_l) \in \mathbb{R}^N$  ▷ Flatten mask vector
5:   $T_l \leftarrow \mathbf{1}_{T_{\text{rep}}} m_l^\top \in \mathbb{R}^{T_{\text{rep}} \times N}$  ▷ Row-wise tiling across $T_{\text{rep}}$ report tokens
6:   $B_l \leftarrow T_l \odot C$  ▷ Apply causal mask
7:   $A'_l \leftarrow A_l + B_l$  ▷ Add anatomical bias to logits
8:   $P_l \leftarrow \text{softmax}(A'_l)$  ▷ Attention weights (used downstream)
9: **end for**

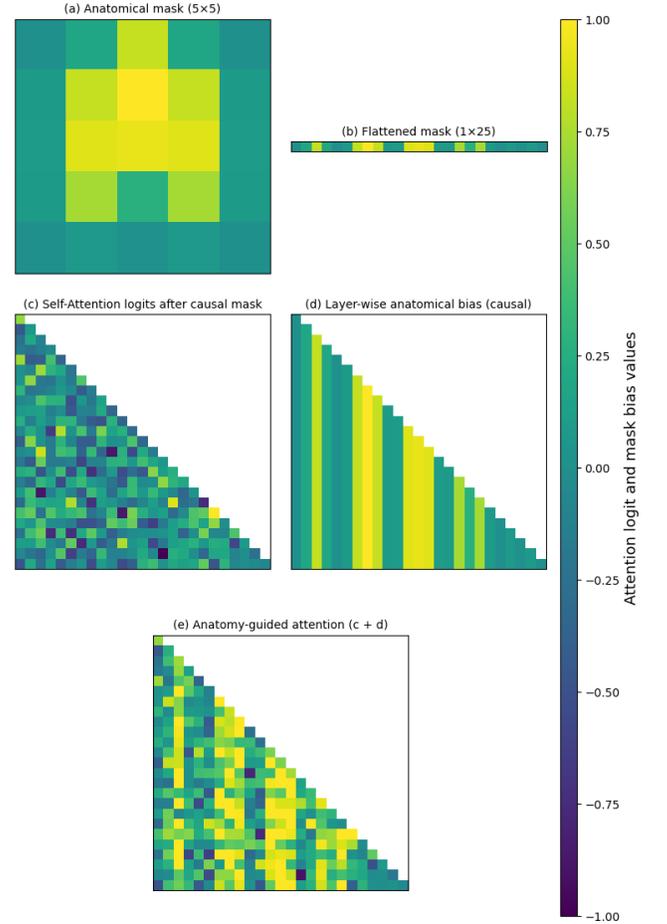

**Fig. 3.** Proposed Layer-wise Anatomical Attention Bias Mechanism. (a) Anatomical mask (simplified to $5 \times 5$ for visualization) representing the Gaussian-smoothed lung-heart regions; (b) Flattened version of the anatomical mask; (c) Self-attention logits after the causal mask; (d) Layer-wise anatomical bias $B_l$ restricted by the causal mask; (e) Anatomy-guided Attention (c + d).

*3) Usage during text generation*

During autoregressive text generation, the sequence length eventually exceeds the number of visual tokens ($N = 1024$). To ensure that anatomical bias influences only the visual portion of the context, the anatomical matrix $T_l$ is extended with zero-padding beyond the first 1024 positions:

- Tokens 1–1024 (visual): receive anatomical bias.
- Tokens 1025 onward (textual): receive zero bias.

This design confines anatomical grounding strictly to the visual part of the attention window, ensuring that as the decoder produces ever longer sequences, the anatomical bias remains targeted and does not interfere with linguistic dependencies in the generated text.

*F. Model architecture*

Figure 4 illustrates the complete architecture of the proposed model, which follows an encoder–adapter–decoder scheme designed to transform a frontal chest radiograph into the *Findings* section of a report through a strictly image-to-text process. This design aligns with modern multimodal architectures where a frozen visual encoder connects to a language model via a linear adaptation module pattern employed in Flamingo [29], BLIP-2 [30], PaLI [31], and LLaVA [32].

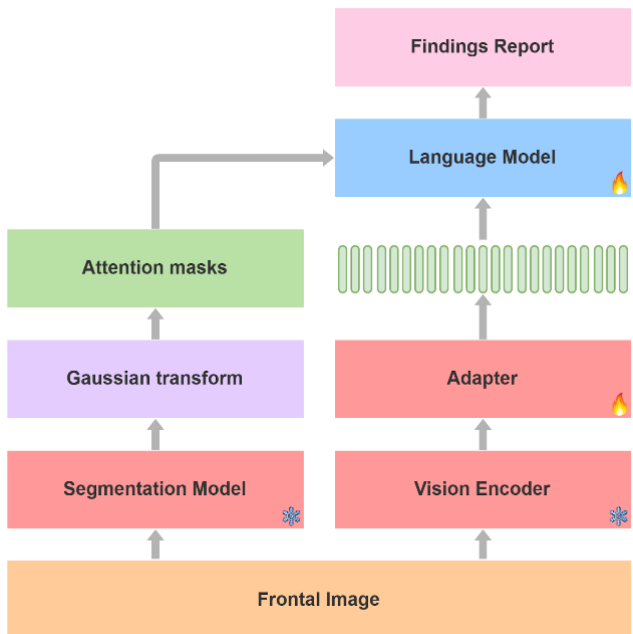

**Fig. 4.** Complete Encoder–Adapter–Decoder Architecture with Explicit Anatomical Attention. Trainable components are indicated by fire icons (🔥), while frozen components are indicated by ice icons (❄).

*1) Visual encoder and adapter*

The visual encoder used is DINOv3 ViT-S/16 [14], an auto-supervised ViT known to extract robust visual representations and shown to generalize well across domains, including medical imaging applications. The encoder receives inputs of size $(B, 3, 512, 512)$ and produces 1024 visual tokens, each of dimension 384, forming a tensor $(B, 1024, 1024)$. This corresponds to a patch embedding partitioning of the frontal radiograph, a standard mechanism in ViT [33] architectures and present in radiological models like MAIRA-1 [7] and MAIRA-2 [8].

The connection between the encoder and the language model is achieved via a multimodal adapter, which applies a linear projection that transforms the visual sequence from $(B, 1024, 384)$ to $(B, 1024, 768)$, matching the embedding dimensionality of GPT-2. This approach is widely used to connect frozen encoders with LLMs, enabling stable training and avoiding modification of millions of encoder parameters [29][30][31][32]. In this work, the adapter is the only trainable module on the visual side.

*2) Textual decoder and context extension*

The textual decoder corresponds to GPT-2 small [15], configured to operate solely as an autoregressive generator conditioned by the visual tokens. Inference is performed without an additional textual prompt: the initial sequence consists only of the projected visual tokens and a start token. To support longer radiological reports, the model's positional capacity was extended using positional interpolation [34], expanding the native embedding length from 1024 to 2048 positions. This technique, originally introduced in ViT [33] and subsequently refined in GPT-NeoX [35], LLaMA [36], ALiBi [28], and RoPE-scaling [37], allows working with long sequences without retraining the model from scratch.

*3) Anatomical Attention Module*

The fundamental component of the system is the anatomical attention module, explicitly shown on the left side of Figure 4. A specialized segmentation model based on ConvNeXt-Small initialized with DINOv3, which generates the binary lung and heart masks. These masks are added directly to the decoder's attention logits before Softmax, providing explicit anatomical grounding without introducing additional trainable parameters. This mechanism is inspired by approaches like MAIRA-SEG [11] and the grounding modules of MAIRA-2 [8], but differs by operating exclusively within the decoder and across all its layers.

Finally, during inference, the model operates in a purely image-to-text mode: the visual encoder and the segmentation model remain frozen, the adapter projects the visual information into the linguistic space, and the decoder generates the report entirely from the image, guided layer-by-layer by the anatomical attention masks.

*G. Training*

Training focused exclusively on the task of generating the *Findings* section from frontal chest radiographs, following the image-to-text paradigm adopted by recent systems such as MAIRA-1 [7] and MAIRA-2 [8]. Throughout the entire process, both the DINOv3 visual encoder [14] and the lung-heart segmentation model remained frozen, aiming to preserve their pre-trained representations and ensure stability when combining a high-capacity Transformer encoder with a lighter linguistic decoder. The only trainable components were the linear adapter, which projects the visual output from dimension 384 to the GPT-2 embedding space of 768, and all layers of the modified linguistic decoder.

Model optimization was performed using causal language modeling, employing an autoregressive cross-entropy loss [15]. During training, the projected visual tokens were concatenated with a start token, and the decoder received the reference text via teacher forcing [8]. No auxiliary tasks or additional signals were added: the decoder was solely trained to model the radiological language distribution of the *Findings* section.

TABLE 2. TRAINING HYPERPARAMETERS

| Hyperparameter | Value |
|---|---|
| Optimizer | AdamW |
| Base Learning Rate | $1 \times 10^{-5}$ |
| Batch Size | 16 |
| Training Epochs | 3 |
| LR Scheduler | Cosine Annealing |
| Warm-up Steps | 5% of total steps |
| Weight Decay | 0.01 |

4The model was trained for 3 epochs on the official MIMIC-CXR-JPG training set. As summarized in Table 2, training utilized the AdamW optimizer [38] with a cosine learning rate scheduler and a 5% warm-up phase. This configuration is consistent with SOTA models for radiological reports, such as MAIRA-2, and was selected to maximize the stability of the decoder without modifying parameters of the visual encoder or segmenter. The entire process was executed on a single 40 GB NVIDIA A100 GPU using Vertex AI within the Google Cloud ecosystem.

*H. Report evaluation*

The model evaluation was conducted exclusively using clinical metrics, selected for their ability to capture the medical quality of the generated content. It has been consistently demonstrated in the literature that traditional linguistic metrics, such as BLEU [39], ROUGE [40], or METEOR [41], do not correlate with the diagnostic utility of radiological reports, as they rely on superficial overlap of textual strings and not on the correct expression of clinical findings [18]. Consequently, the best-performing reference models, including MAIRA-1 [7], MAIRA-2 [8] and MedPaLM-M [9], have adopted exclusively structured medical metrics. This work follows the same standard, evaluating the model quality only through methods that directly measure the diagnostic and anatomical coherence of the report.

*1) CheXpert Metrics*

To evaluate the clinical accuracy of the findings, CheXbert was used, a BERT-based model designed to automatically extract the 14 pathologies defined by the CheXpert scheme [2][17]. This set includes No Finding, Support Devices, Fracture, Enlarged Cardiomegaly, Cardiomegaly, Pleural Other, Pleural Effusion, Lung Opacity, Lung Lesion, Edema, Consolidation, Pneumonia, Atelectasis, and Pneumothorax. CheXbert assigns normalized labels for presence, absence, or uncertainty, which are subsequently binarized using standard protocols described in the literature. Two variants of the clinical F1 score are reported:
- Macro/Micro F1 over the 14 pathologies.
- Macro/Micro F1 over five clinically dominant pathologies (cardiomegaly, edema, consolidation, pleural effusion, and atelectasis).

These metrics are considered fundamental in the evaluation of report generation models because they quantify clinical precision rather than superficial textual coincidence [18].

*2) RadGraph Metrics*

The semantic structure of the generated report was evaluated using RadGraph, an extraction system that converts radiological reports into clinical graphs composed of anatomical entities, findings, and semantic relations between them [18]. This representation allows for quantifying whether the model correctly identifies chest structures, pathologies, and their spatial or logical relationships.

- The RadGraph F1 metric, employed in recent benchmarks like MAIRA-1, MAIRA-2, and RadCliQ, measures the simultaneous correctness of entities and relations, and is considered the current standard for evaluating clinical grounding.
- The RadGraph-ER variant, used in prior works like R2Gen [6], evaluates the correctness of the span and the clinical type, provided the entity participates in at least one valid relation. This metric is less strict but useful for historical comparability and for analyzing models with varying grounding capabilities.

We performed all CheXpert and RadGraph calculations using the implementations described in [42][43]. The evaluation was carried out exclusively on the official MIMIC-CXR-JPG test set [16], after filtering studies containing the *Findings* section, following the protocol used by MAIRA-1 [7], MAIRA-2 [8], and MedPaLM-M [9]. This procedure guarantees direct comparability with the SOTA and ensures that the results reflect the real diagnostic quality of the generated text, avoiding the inherent limitations of purely linguistic metrics and prioritizing those that represent clinical meaning.

## III. RESULTS

The lung segmentation model evaluation on the test set yielded a Dice score of 0.9394, an IoU of 0.8944, a sensitivity of 0.9343, a specificity of 0.9757, and an overall accuracy of 0.9639, consistent with SOTA segmentation methods. The heart segmentation model achieved a Dice score of 0.9472, an IoU of 0.9036, an accuracy of 0.9742, a sensitivity of 0.9662, and a specificity of 0.9779. These results confirm that the model successfully captured the lungs and cardiac silhouette despite anatomical variability and common radiographic confounders such as mediastinal widening.

The experimental evaluation of the anatomical mask was performed using three model variants (NO-MASK, MASK, and HIDDEN-MASK) trained with the DINOv3 visual encoder [14] frozen, yielding a total of 3 models. Each configuration was trained with 8,000 studies and approximately 600,000 tokens, allowing for a controlled analysis of the effect of anatomical bias incorporated into the decoder and the impact of freezing the encoder. The MASK variant corresponds to the proposed method, where the smoothed anatomical mask is directly added to the attention logits. HIDDEN-MASK uses the same mask, but tokens where the mask is zero receive a value of $-\infty$, eliminating them from attention. Finally, NO-MASK represents the baseline configuration without added anatomical information. These three variants allow us to isolate the benefit of explicit anatomical bias and compare it against alternatives without such a structure.

*A. Quantitative Comparison*

Initial results consistently show that the explicit incorporation of the anatomical mask significantly increases the clinical quality of the generated report. Figure 5 presents the quantitative comparison between the three strategies, where the MASK variant is observed to outperform NO-MASK and HIDDEN-MASK in all CheXpert F1 metrics and most RadGraph metrics, with increments ranging from 9.7% up to 186.4% depending on the metric and the encoder state. This behavior confirms that the anatomical bias incorporated into the decoder directs attention towards medically relevant regions, favoring the production of clinical tokens instead of purely linguistic sequences lacking diagnostic meaning.

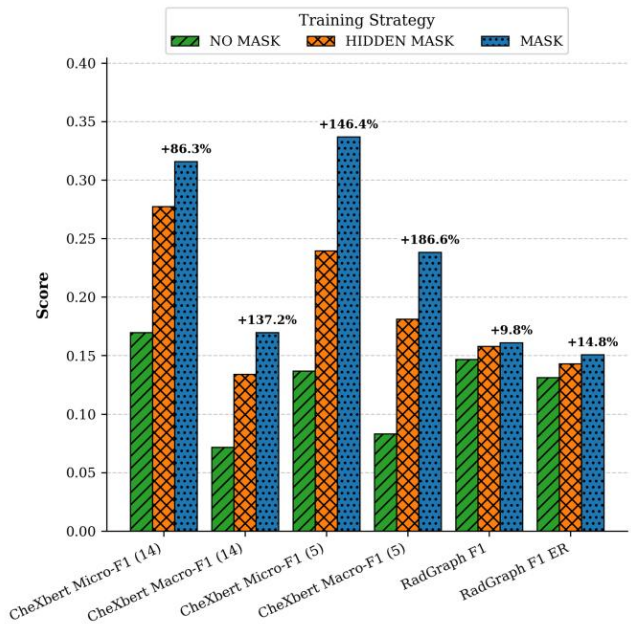

**Fig. 5.** Quantitative Comparison of Clinical Performance under Frozen DINOv3 Encoder.

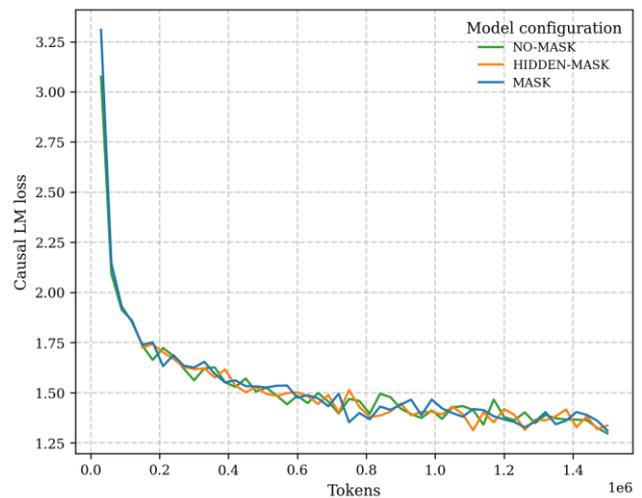

**Fig. 6.** Causal Language Model Loss Evolution Across All Three Evaluated Configurations.

Figure 6 shows the evolution of Causal Language Model (LM) Loss during training. Although the models exhibit indistinguishable loss curves, this metric only reflects the model's ability to predict the next token, without distinguishing between clinically relevant tokens or semantically empty text. Therefore, the linguistic loss does not adequately predict medical quality, justifying the need for clinical metrics like CheXbert F1 and RadGraph to evaluate the real impact of anatomical bias.

From the experiments, the MASK strategy produced consistent and substantial improvements across all clinical metrics when compared with the NO MASK baseline. CheXpert Micro-F1 (14) rose from 0.1696 to 0.3159 (86.26% increase) and Macro-F1 (14) improved from 0.0715 to 0.16.97 (137.34% increase). The largest relative improvements were observed for the most clinically relevant subset: Micro-F1 (5) increased from 0.13.68 to 0.337 (146.35%), and Macro-F1 (5) from 0.0832 to 0.2383 (168.42%). Structural performance metrics also improved: RadGraph F1 rose from 0.1466 to 0.1609 (9.75%), and RadGraph-ER from 0.1314 to 0.1509 (14.84%). Together, these results show that anatomical attention in the decoder provides strong clinical benefits, particularly for capturing common and subtle thoracic findings.

### B. Findings Generation

The final training was carried out exclusively with MASK implementation on the MIMIC-CXR-JPG training set [16]. On the official MIMIC-CXR-JPG test split for the *Findings* section generation, the model achieved CheXpert Micro-F1 and Macro-F1 scores of 0.323 and 0.219, respectively, for the 14-label set. For the 5-label subset, the model recorded a Micro-F1 of 0.374 and a Macro-F1 of 0.335. Regarding RadGraph metrics, the model yielded an F1 score of 0.183 and an Entity and Relation (ER) score of 0.159.

### IV. DISCUSSION

The results obtained demonstrate that a compact model based on a frozen DINOv3 encoder [14] and a modified GPT-2 decoder [15] can generate radiological reports with reasonable clinical coherence, even without resorting to massive multimodal architectures like MAIRA-2 [8] (7B parameters), MedPalM-M [9], or multimodal systems trained with extensive clinical data. Our complete model, with a total of approximately 245M parameters (21M DINOv3 Encoder + 2 ×50M ConvNeXt-Small Segmenter + 124M GPT-2 Decoder), is significantly smaller than these leading models.

The observed gap compared to the SOTA is expected, given that current leading models not only employ larger encoders and decoders with billions of parameters but also integrate multiple sources of clinical grounding such as lateral views, previous reports, prior studies, structured metadata, and clinical history, whereas the model presented in this work operates exclusively with a single frontal chest X-ray as input.

A critical distinction of our approach is the lack of an initial prompt, forcing the GPT-2 decoder to rely strictly on the learned distribution from the training data. This design decision, while simplifying the input pipeline, makes the model highly sensitive to the nature and completeness of the reports used for training. Specifically, we observed that the model's scope is inherently constrained by the quality and content of expert reports, which may omit visually apparent findings that radiologists judge as clinically irrelevant in light of the patient's

4history, a phenomenon well documented in prior studies on contextual and comparison bias in chest radiograph interpretation [44][45]. Thus, not all visible characteristics in the X-ray are encoded in the textual output, but only those considered clinically pertinent, establishing an unavoidable ceiling on the model's descriptive potential.

*A. Validation of anatomic attention*

Despite these structural limitations, the observed performance validates the central hypothesis of this work: explicit anatomical attention within the decoder improves the spatial and semantic coherence of the generated text. The layer-wise anatomical mask mechanism, derived from lung and heart segmentation, produces substantial clinical improvements in controlled scenarios.

In brief training sessions of 8,000 studies, the use of anatomical attention (MASK) yielded considerable clinical increases over the model without anatomical bias (NO-MASK): in CheXpert Macro-F1(5), Micro-F1(5), Macro-F1(14) [17], and more moderate improvements in RadGraph F1 [18]. We hypothesize that these results could stem from two main effects of the proposed method. First, the explicit guidance may reduce the training time required for convergence, effectively making the model faster to train by narrowing the search space. Second, the masking mechanism potentially improves the signal-to-noise ratio: it appears to add necessary anatomic information while simultaneously reducing unneeded visual information, thereby helping the model focus on what is clinically pertinent. These increments suggest that the model, by receiving hierarchical anatomical information, prioritizes clinically relevant tokens over irrelevant textual descriptions, reducing hallucinations and improving semantic accuracy.

A notable finding is that this mechanism particularly benefits models designed to mimic the way a radiologist checks an X-ray, specifically those that focus on constant anatomical regions, such as the lungs and heart. For these specialized models (or unimodal models focused on a single task), the incorporation of anatomical attention can act as a strong inductive bias that facilitates spatial grounding. In contrast, for multi-task models that require exploring different regions depending on the task (e.g., bone fractures, abdomen, neck, or non-anatomical classification tasks), this bias could become counterproductive by excessively restricting the attention distribution. Therefore, the proposed technique is especially suitable for compact, focused, and single-task models, or for architectures where anatomical information is stable across tasks.

*B. SOTA comparison*

The results obtained were compared against the current SOTA in automatic radiological report generation. Table 3 summarizes the performance against models like MAIRA-2 [8] and MedPaLM-M [9]. Although our system does not reach the absolute values of these large-scale models (which employ deeper architectures, higher parameter counts, clinical prompts, multimodal reasoning, and intensive human supervision) it positions itself as a compact and clinically competent model within its category.

The results demonstrate that integrating anatomical information directly into the decoder clearly improves the semantic and anatomical coherence of the generated report, even while keeping the visual encoder frozen and using only a linear adapter. The model produces clinically reasonable reports, reducing hallucinations outside the relevant anatomical regions and favoring the generation of descriptions more congruent with the lungs and heart, thereby validating the central hypothesis of this work: incorporating anatomical attention into the decoder is an effective and efficient way to improve image-text alignment without relying on massive models or complex architectures.

TABLE 3. COMPARISON OF CLINICAL PERFORMANCE: PROPOSED MODEL (MASK + FROZEN) VS. SOTA ON THE MIMIC-CXR OFFICIAL TEST SPLIT

| METRIC | SOTA | PROPOSED MODEL |
|---|---|---|
| CHEXPERT MICRO-F1 (14) | **0.581** (MAIRA-2) | 0.323 |
| CHEXPERT MACRO-F1 (14) | **0.416** (MAIRA-2) | 0.219 |
| CHEXPERT MICRO-F1 (5) | **0.591** (MAIRA-2) | 0.374 |
| CHEXPERT MACRO-F1 (5) | **0.516** (MEDPALM-M) | 0.335 |
| RADGRAPH F1 | **0.346** (MAIRA-2) | 0.183 |
| RADGRAPH ER | **0.396** (MAIRA-2) | 0.159 |

*C. Hierarchical Smoothing and Future Work*

The use of smoothed masks via Gaussian filters also proved beneficial. Following a hierarchical "general-to-specific" approach, the initial decoder layers receive broad masks, while the top layers obtain more focused masks. This design allowed the model to maintain a global understanding of the chest in early stages and refine its attention toward the lungs and heart as generation progressed, achieving a more precise anatomical correspondence without adding trainable parameters.

Although the final model does not achieve the clinical performance of MAIRA-2 or MedPaLM-M, this work contributes a new technique for anatomical bias directly in the decoder to the community, demonstrating that even small models [29] can benefit from incorporating explicit anatomical structures within the self-attention mechanism. Resource limitations prevented the exploration of larger variants, prolonged training, or models with clinically specialized encoders; however, the presented evidence suggests that the technique can be scaled and potentially combined with more powerful architectures. In this way, this work introduces a significant contribution toward the efficient integration of anatomical knowledge in clinical generative models and opens a concrete path for future research in anatomical grounding within language models applied to medical imaging.

CONCLUSION

We presented a compact and resource efficient image to text architecture for chest X ray report generation that incorporates explicit anatomical guidance at the decoder level. While the proposed approach does not outperform large scale SOTA systems, it demonstrates that structured spatial priors derived from lung and heart segmentation masks can significantly improve clinically relevant performance in purely image conditioned settings.

Layer wise anatomical attention injected into a lightweight GPT 2 decoder improved both clinical accuracy and structural coherence, as reflected by gains in CheXpert and RadGraph metrics. These results suggest that guiding attention toward regions of interest reduces the influence of irrelevant visual information and enhances spatial grounding during report generation.

Although developed for chest X-ray report generation, the proposed mechanism is model and task independent. It could be applied to larger language vision architectures, extended to other medical imaging modalities, or transferred to non-medical image to text tasks where salient spatial regions can be estimated.

Additional experimentation is required to clarify the origin of the observed performance gains, specifically whether they result from improved optimization behavior or from a true enhancement in representational quality due to reduced attention to irrelevant image regions. Overall, this work positions decoder level region guided attention as a practical and scalable approach for improving image to text generation without increasing model size or training complexity.

ACKNOWLEDGMENTS

This work was supported by computational resources provided by Google Cloud. The authors also acknowledge the essential contribution of public medical datasets to this research. We thank the creators and maintainers of the CheXpert, MIMIC-CXR, DARWIN, Montgomery County, Shenzhen, and JSRT datasets for their service to the scientific community.